\documentclass[twoside,twocolumn]{article}

\usepackage[scaled=.95]{newtxtext,newtxmath} 
\usepackage[T1]{fontenc} 
\usepackage{microtype} 

\usepackage[polish,slovene,english]{babel} 
\usepackage[utf8]{inputenc} 

\usepackage{graphicx} 
\usepackage{makecell} 
\usepackage{tabularx} 
\usepackage{booktabs}
\usepackage{bchart} 
\usepackage{multirow} 
\usepackage[title]{appendix}

\usepackage[hang]{footmisc} 
\usepackage[hmarginratio=1:1,margin={1.9cm,1.9cm},top=26mm,bottom=26mm,columnsep=20pt]{geometry}
\usepackage[hang, small,labelfont=bf,up,textfont=bf,up,format=plain]{caption} 
\usepackage{booktabs} 

\usepackage{lettrine} 

\usepackage{enumitem} 
\setlist[itemize]{noitemsep} 

\usepackage{abstract} 

\usepackage{titlesec} 
\titleformat{\section}[block]{\large\bfseries}{\thesection}{1em}{\MakeUppercase}{} 
\titlespacing*\section{0pt}{6pt}{5pt}
\titleformat{\subsection}[block]{\large\bfseries}{\thesubsection}{1em}{} 
\titlespacing*\subsection{0pt}{6pt}{5pt}
\titleformat{\subsubsection}[runin]{\normalsize\itshape}{\thesubsubsection}{1em}{} \titlespacing*\subsubsection{0pt}{6pt}{5pt}

\usepackage{fancyhdr} 
\fancypagestyle{specialfooter}{%
  \fancyhf{}
  
  \fancyfoot[L]{©2022 Copyright held by the authors. This is the authors' version of the work. It is posted here for your personal use. Not for redistribution. The definitive version will be published in the Proceedings of the 2023 IEEE/CVF Winter Conference on Applications of Computer Vision (WACV).}
}

\usepackage{titling} 
\usepackage{hyperref} 

\usepackage[round]{natbib} 
\newlength{\bibitemsep}
\newlength{\bibparskip}
\let\oldthebibliography\thebibliography
\renewcommand\thebibliography[1]{%
  \oldthebibliography{#1}%
  \setlength{\parskip}{\bibitemsep}%
  \setlength{\itemsep}{\bibparskip}%
}

\usepackage{upgreek}
\newcommand\omicron{o}
\newcommand\Omicron{O}
\newcommand{\Layer}{\mathcal{L}} 
\newcommand{\Msk}{\mathcal{M}} 
\newcommand{\Ele}{\mathcal{E}} 
\newcommand{\dbar}{d\hspace*{-0.08em}\bar{}\hspace*{0.1em}}
\usepackage{textcomp}
\usepackage{multicol}
\usepackage{makecell}
\usepackage[flushleft]{threeparttable}
\usepackage{algorithm,algorithmic}

\title{A Priority Map for Vision-and-Language Navigation with Trajectory Plans and Feature-Location Cues} 
\author{%
\textsc{Jason Armitage} \\
\normalsize University of Zurich \\
\normalsize Switzerland
\and 
\textsc{Leonardo Impett} \\
\normalsize University of Cambridge \\
\normalsize UK
\and 
\textsc{Rico Sennrich} \\
\normalsize University of Zurich \\
\normalsize Switzerland
}
\date{} 

\begin{document}

\maketitle
\pagestyle{empty} 
\thispagestyle{specialfooter}

\begin{abstract}
\vspace*{-.9em}
\noindent 
In a busy city street, a pedestrian surrounded by distractions can pick out a single sign if it is relevant to their route. Artificial agents in outdoor Vision-and-Language Navigation (VLN) are also confronted with detecting supervisory signal on environment features and location in inputs. To boost the prominence of relevant features in transformer-based systems without costly preprocessing and pretraining, we take inspiration from priority maps - a mechanism described in neuropsychological studies. We implement a novel priority map module and pretrain on auxiliary tasks using low-sample datasets with high-level representations of routes and environment-related references to urban features. A hierarchical process of trajectory planning - with subsequent parameterised visual boost filtering on visual inputs and prediction of corresponding textual spans - addresses the core challenge of cross-modal alignment and feature-level localisation. The priority map module is integrated into a feature-location framework that doubles the task completion rates of standalone transformers and attains state-of-the-art performance for transformer-based systems on the Touchdown benchmark for VLN. \href{https://github.com/JasonArmitage-res/PM-VLN}{Code} and \href{https://zenodo.org/record/6891965#.YtwoS3ZBxD8}{data} are referenced in Appendix \ref{app_c}.
\end{abstract}


\section{Introduction}

Navigation in the world depends on attending to relevant cues at the right time. A road user in an urban environment is presented with billboards, moving traffic, and other people - but at an intersection will pinpoint a single light to check if it contains the colour red~\citep{gottlieb2020curiosity, shinoda2001controls}. An artificial agent navigating a virtual environment of an outdoor location is also presented with a stream of linguistic and visual cues. Action selections that move the agent closer to a final destination depend on the prioritisation of references that are relevant to the point in the trajectory. In the first example, human attention is guided to specific objects by visibility and the present objective of crossing the road. At a neurophysiological level, this process is mediated by a priority map - a neural mechanism that guides attention by matching low-level signals on salient objects with high-level signals on task goals. Prioritisation in humans is enhanced by combining multimodal signals and integration between linguistic and visual information~\citep{ptak2012frontoparietal, cavicchio2014effect}. The ability to prioritise improves as experience of situations and knowledge of environments increases~\citep{zelinsky2015and, tatler2011eye}. 

We introduce a priority map module for Vision-and-Language Navigation (PM-VLN) that is pretrained to guide a transformer-based architecture to prioritise relevant information for action selections in navigation. In contrast to pretraining on large-scale datasets with generic image-text pairs~\citep{DBLP:conf/iclr/SuZCLLWD20}, the PM-VLN module learns from small sets of samples representing trajectory plans and urban features. Our proposal is founded on observation of concentrations in location deictic terms and references to objects with high visual salience in inputs for VLN. Prominent features in the environment pervade human-generated language navigation instructions. Road network types (“intersection”), architectural features (“awning”), and transportation (“cars”) all appear with high frequency in linguistic descriptions of the visual appearance of urban locations. Learning to combine information in the two modalities relies on synchronising temporal sequences of varying lengths. We utilise references to entities as a signal for a process of cross-modal prioritisation that addresses this requirement.

Our module learns over both modalities to prioritise timely information and assist both generic vision-and-language and custom VLN transformer-based architectures to complete routes~\citep{li2019visualbert, zhu2021multimodal}. Transformers have contributed to recent proposals to conduct VLN, Visual Question Answering, and other multimodal tasks - but are associated with three challenges: 1) Standard architectures lack mechanisms that address the challenge of temporal synchronisation over linguistic and visual inputs. Pretrained transformers perform well in tasks on image-text pairs but are challenged when learning over sequences without explicit alignments between modalities~\citep{lin2020audiovisual}. 2) Performance is dependent on pretraining with large sets of image-text pairs and a consequent requirement for access to enterprise-scale computational resources~\citep{majumdar2020improving, suglia2021embodied}. 
3) Visual learning relies on external models and pipelines - notably for object detection~\citep{li2020oscar, le2022object}. The efficacy of object detection for VLN is low in cases where training data only refer to a small subset of object types observed in navigation environments.

We address these challenges with a hierarchical process of trajectory planning with feature-level localisation and low-sample pretraining on in-domain data. We use discriminative training on two auxiliary tasks that adapt parameters of the PM-VLN for the specific challenges presented by navigating routes in outdoor environments. High-level planning for routes is enabled by pretraining for trajectory estimation on simple path traces ahead of a second task comprising multi-objective cross-modal matching and location estimation on urban landmarks. Data in the final evaluation task represent locations and trajectories in large US cities and present an option to leverage real-world resources in pretraining. Our approach builds on this opportunity by sourcing text, images, coordinates, and path traces from the open web and the Google Directions API where additional samples may be secured at low cost in comparison to human generation of instructions.

\begin{figure*}
\begin{center}
\centering
\captionsetup{type=figure}
\includegraphics[width=160mm,scale=0.5]{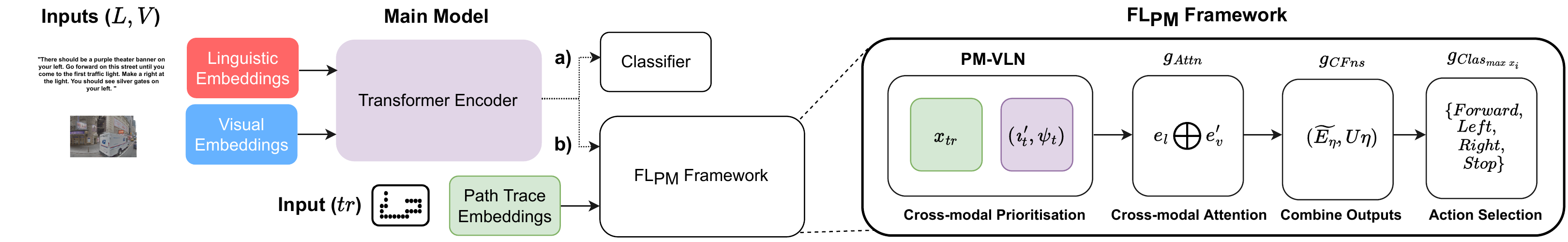}
\captionof{figure}{Prior work on transformer-based systems for VLN follows the above pipeline from inputs to the main model concluding with a) a classifier to predict actions. We propose a feature-location framework (FL\textsubscript{PM}) to enhance the performance of a main model as in b). Here path traces are an additional input to assist the PM-VLN to align linguistic and visual sequences. Submodule $g_{CFns}$ combines embeddings from the main model $U_{\eta}$ and the PM-VLN $\widetilde{E_\eta}$ ahead of action prediction with maxout activation.}
\label{flpm}
\end{center}
\vspace{-4mm} 
\end{figure*}

\begin{figure}[hbt!]
\begin{center}
\centerline{\includegraphics[width=\columnwidth]{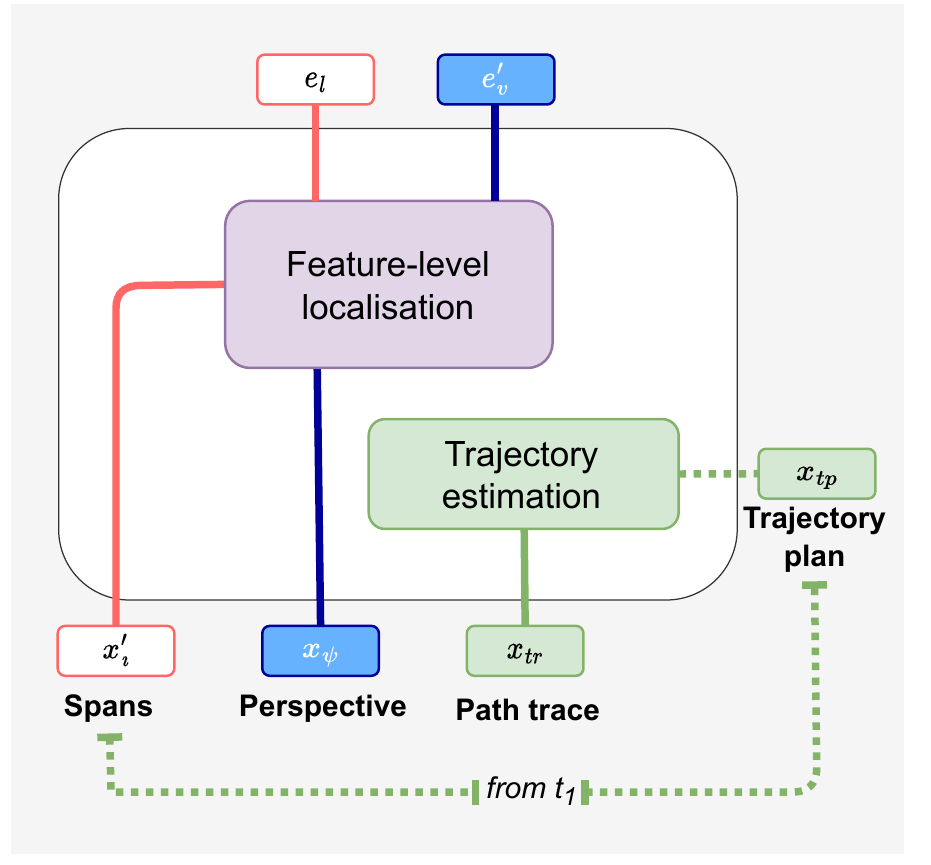}}
\caption{A Priority Map module performs a hierarchical process of high-level trajectory planning and feature-level localisation. Submodules inside the white box are learned together and a helper function generates a trajectory plan to predict spans from step $t_1$.}
\label{figpmrev2}
\end{center}
\vskip -0.3in
\end{figure}

This research presents four contributions to enhance transformer-based systems on outdoor VLN tasks:
\begin{itemize}[topsep=0pt]
\item \textbf{Priority map module} Our novel PM-VLN module conducts a hierarchical process of high-level alignment of textual spans with visual perspectives and feature-level operations to enhance and localise inputs during navigation (see Figure \hyperref[figpmrev2]{3}).
\item \textbf{Trajectory planning} We propose a new method for aligning temporal sequences in VLN comprising trajectory estimation on path traces and subsequent predictions for the distribution of linguistic descriptions over routes.
\item \textbf{Two in-domain datasets and training strategy} We introduce a set of path traces for routes in two urban locations (TR-NY-PIT-central) and a dataset consisting of textual summaries, images, and World Geodetic System (WGS) coordinates for landmarks in 10 US cities (MC-10). These resources enable discriminative training of specific components of the PM-VLN on trajectory estimation and multi-objective loss for a new task that pairs location estimation with cross-modal sentence prediction.
\item \textbf{Feature-location framework} We design and build a framework (see Figure \hyperref[flpm]{2}) to combine the outputs from the PM-VLN module and cross-modal embeddings from a transformer-based encoder. The framework incorporates components for performing self-attention, combining embeddings, and predicting actions with maxout activation.
\end{itemize}


\section{Background} \label{background}
In this section we define the Touchdown task and highlight a preceding challenge of aligning and localising over linguistic and visual inputs addressed in our research. A summary of the notation used below and in subsequent sections is presented in Appendix \ref{app_a}.

\textbf{Touchdown} Navigation in the Touchdown benchmark $\phi_{VLN}$ is measured as the completion of $N$ predefined trajectories by an agent in an environment representing an area of central Manhattan. The environment is represented as an undirected graph composed of nodes $\Omicron$ located at WGS latitude / longitude points. At each step $t$ of the sequence $\{1, …, T\}$ that constitute a trajectory, the agent selects an edge $\xi_t$ to a corresponding node. The agent’s selection is based on linguistic and visual inputs. A textual instruction $\uptau$ composed of a varying number of tokens describes the overall trajectory. We use $\varsigma$ to denote a span of tokens from $\uptau$ that corresponds to the agent’s location in the trajectory. Depending on the approach, $\varsigma$ can be the complete instruction or a selected sequence. The visual representation of a node in the environment is a panorama drawn from a sequence $Route$ of undetermined length. The agent receives a specific perspective $\psi$ of a panorama determined by the heading angle $\angle$ between $(\omicron_1, \omicron_2)$. Success in completing a route is defined as predicting a path that ends at the node designated as the goal - or one directly adjacent to it.

\begin{figure}[hbt!]
\vskip -0.1in
\begin{center}
\centerline{\includegraphics[width=\columnwidth]{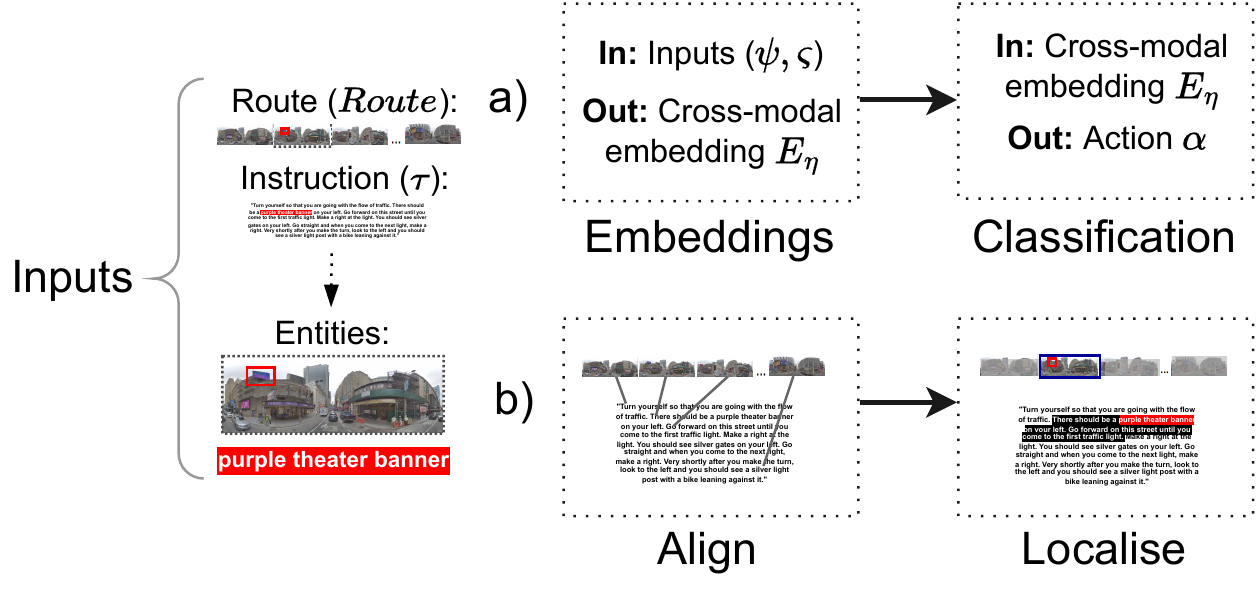}}
\caption{Outline of VLN as a supervised classification task a). Linguistic and visual inputs both refer to entities indicated in red. We address a challenge to align and localise over unsynchronised inputs b) by focusing on entities represented in both modalities.}
\label{task}
\end{center}
\vskip -0.3in
\end{figure}

In a supervised learning paradigm (see a) in Figure \hyperref[task]{1}), an embedding $e_\eta$ is learned from inputs $\varsigma_t$ and $\psi_t$. The agent’s next action is a classification over $e_\eta$ where the action $\alpha_t$ is one of a class drawn from the set $\mathrm{A}\{Forward, Left, Right, Stop\}$. Predictions $\alpha_t = Forward$ and $\alpha_t = \{Left, Right\}$ result respectively in a congruent or a new $\angle$ at edge $\xi_{t+1}$. A route in progress is terminated by a prediction $\alpha_t = Stop$. 

\textbf{Align and Localise} We highlight in Figure \hyperref[task]{1} a preceding challenge in learning cross-modal embeddings. As in real-world navigation, an agent is required to align and match cues in instructions with its surrounding environment. A strategy in human navigation is to use entities or landmarks to perform this alignment~\cite{cavicchio2014effect}. In the Touchdown benchmark, a relationship between sequences $\uptau$ and $Route$ is assumed from the task generation process outlined in \cite{chen2019touchdown} - but the precise alignment is not known. We define the challenge as one of aligning temporal sequences $\uptau = \{\varsigma_1, \varsigma_2, …, \varsigma_n\}$ and $Route = \{\psi_1, \psi_2, …, \psi_n\}$ with the aim of generating a set of cross-modal embeddings $E_\eta$ where referenced entities correspond. At a high level, this challenge can be addressed by an algorithm $q$ that maximises the probability $P$ of detecting $S$ signal in a set of inputs. This algorithm is defined as 
\begin{equation}
q(\theta)((X_t)) = q(\theta)(X_t) \rightarrow max \left[\int_{\mathcal{X}} p(X_t|\theta) s(X_t) \right]
\end{equation}
where $\theta$ is a parameter $\theta \in \Theta_1$ and $\mathcal{X}$ is the data space.

In the Touchdown benchmark, linguistic and visual inputs are of the form $0 \leq |\uptau| \leq n$ and $0 \leq |Route| \leq n$ where $len(\uptau) \neq len(Route)$. The task then is to maximise the probability of detecting signal in the form of corresponding entities over the sequences $\uptau$ and $Route$, which in turn is the product of probabilities over pairings $\varsigma_t$ and $\psi_t$ presented at each step:
\begin{equation}
\underset{subject \, to}{g(X_t) \rightarrow max} \; \; P[\uptau, Route] = \prod p_{x_\varsigma x_\psi} 
\end{equation}


\section{Method}
We address the challenge of aligning and localising over sequences with a computational implementation of cross-modal prioritisation. Diagnostics on VLN systems have placed in question the ability of agents to perform cross-modal alignment~\citep{zhu-etal-2022-diagnosing}. Transformers underperform in problems with temporal inputs where supervision on image-text alignments is lacking~\citep{chen2020fine}. This is demonstrated in the case of Touchdown where transformer-based systems complete less than a quarter of routes. Our own observations of \hyperref[table:res1]{lower performance} when increasing the depth of transformer architectures motivates moving beyond stacking blocks to an approach that compliments self-attention.

Our PM-VLN module modulates transformer-based encoder embeddings in the main task $\phi_{VLN}$ using a hierarchical process of operations and leveraging prior learning on auxiliary tasks $(\phi_1, \phi_2)$ (see Figure \hyperref[pm_module]{3}). In order to prioritise relevant information, a training strategy for PM-VLN components is designed where training data contain samples that correspond to the urban grid type and environment features in the main task. The datasets required for pretraining contain less samples than other transformer-based VLN frameworks~\citep{zhu2021multimodal, majumdar2020improving} and target only specific layers of the PM-VLN module. The pretrained module is integrated in a novel feature-location framework FL\textsubscript{PM} shown in Figure \hyperref[flpm]{2}. Subsequent components in the FL\textsubscript{PM} combine cross-modal embeddings from the PM-VLN and a main transformer model ahead of predicting an action.

\subsection{Feature-location Framework with a Priority Map Module}
Prior work on VLN agents has demonstrated reliance for navigation decisions on environment features and location-related references~\citep{zhu2021multimodal}. In the definition of $\phi_{VLN}$ \hyperref[background]{above}, we consider this information as the supervisory signal contained in both sets of inputs $(x_\varsigma, x_\psi)_t$. As illustrated in Figure \hyperref[flpm]{2}, our PM-VLN module is introduced into a framework FL\textsubscript{PM}. This framework takes outputs from a transformer-based main model $Enc_{Trans}$ together with path traces ahead of cross-modal prioritisation and classification with maxout activation $Clas_{max \ x_i}$. Inputs for $Enc_{Trans}$ comprise cross-modal embeddings $\bar{e}_{\eta}$ proposed by \cite{zhu2021multimodal} and a concatenation of perspectives up to the current step $\psi_{cat}$
\begin{equation}
\begin{split}
Clas_{\underset{i}{max \, x_i}}[y_j | z'] &= \dbar(\textit{PM-VLN}(\{g(x_i), (tr_t, \imath_t', \psi_t)\}_{i=1}^n) + \\ & Enc_{Trans}(\{g(x_i), (\bar{e}_{\eta}, \psi_{cat})\}_{i=1}^n))
\end{split}
\end{equation} 
where $tr_t$ is a path trace, $z'$ is the concatenation of the outputs of the two encoders, and $\dbar$ is a dropout operation.

\subsubsection{Priority Map Module}
Priority maps are described in the neuropsychological literature as a mechanism that modulates sensory processing on cues from the environment. Salience deriving from the physical aspects of objects in low-level processing is mediated by high-level signals for the relevance of cues to task goals~\citep{fecteau2006salience, itti2000saliency, zelinsky2015and}. Cortical regions that form the location of these mechanisms are associated with the combined processing of feature- and location-based information~\citep{bisley2019neural, hayden2009combined}. Prioritisation of items in map tasks with language instructions indicate an integration between linguistic and visual information and subsequent increases in salience attributed to landmarks~\citep{cavicchio2014effect}.

Our priority map module (PM-VLN) uses a series of simple operations to approximate the prioritsation process observed in human navigation. These operations avoid dependence on initial tasks such as object detection.  Alignment of linguistic and visual inputs is enabled by trajectory estimation on simple path traces forming high-level representations of routes and subsequent generation of trajectory plans. Localisation consists of parameterised visual boost filtering on the current environment perspective $\psi_t$ and cross-modal alignment of this view with selected spans from subsequent alignment (see Algorithm \hyperref[alg:pmm]{1}). This hierarchical process compliments self-attention by accounting for the lack of a mechanism in transformers to learn over unaligned temporal sequences. A theoretical basis for cross-modal prioritisation is presented \hyperref[theo]{below}.

\begin{algorithm}[H]
\caption{Priority Map Module}
\label{alg:pmm}
\begin{algorithmic}
\STATE {\bfseries Input:} Datasets $\mathcal{D}_{\phi_1}$,$\mathcal{D}_{\phi_2}$, and $\mathcal{D}_{\phi_{VLN}}$ with inputs $(x_l, x_v)$ for tasks $\Phi$. Initial parameters in all layers at $\Theta_j^l\sim Normal(\mu_j,\sigma_j).$
\STATE {\bfseries Output:} $(e_l, e_v')$
\WHILE{not converged}
\FOR{$x_{{tr}_i}$ {\bfseries in} $\phi_1$}
    \STATE $\Theta_{g_{PMTP}}' \leftarrow g_{\phi_1}(X_i, \Theta).$
    \ENDFOR
    \ENDWHILE
    \WHILE{not converged}
    \FOR{$(x_{l_i},x_{v_i})$ {\bfseries in} $\phi_2$}
        \STATE $\Theta_{g_{PMF}}' \leftarrow g_{\phi_2}(X_i, \Theta).$
        \ENDFOR
        \ENDWHILE
        \WHILE{not converged}
        \STATE Sample $x_{tr_t}$ from $D^{Train}.$
        \STATE $x_{tp_t}  \leftarrow{g_{PMTP}(x_{tr_t})}.$
        \STATE Sample $(\imath_t', \psi_t)$ from $D^{Train}.$
        \STATE $e_v \leftarrow{g_{USM}(\psi_t)}.$
        \STATE $e_v' \leftarrow{g_{VBF}(e_v)}.$
        \STATE $e_l \leftarrow{g_{PrL}(g_{Cat}(\imath_t', e_v'))}.$
        \ENDWHILE
\STATE {\bfseries return} $(e_l, e_v')$
\end{algorithmic}
\end{algorithm}

\begin{figure}[hbt!]
\begin{center}
\centerline{\includegraphics[width=\columnwidth]{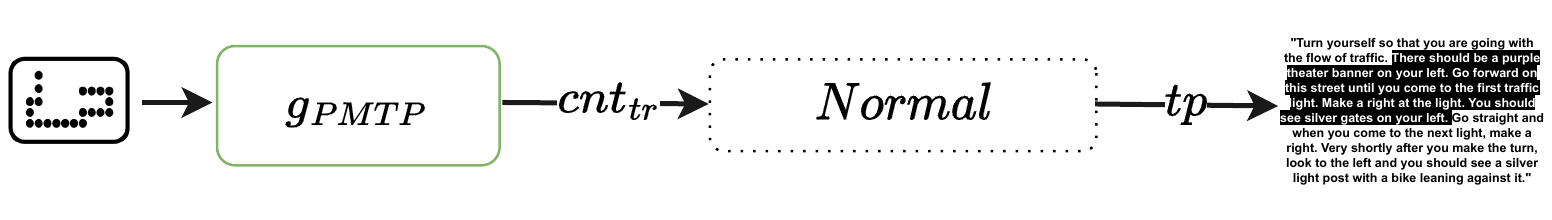}}
\caption{Submodule $g_{PMTP}$ estimates a step count ($cnt_{tr}$) on a path trace. A trajectory plan ($tp$) is a Gaussian distribution ($Normal$) over the instruction and predicts a span for every step  $\imath_t$. This is concatenated with the span predicted for the previous step.}
\label{pmtp}
\end{center}
\vskip -0.3in
\end{figure}

\textbf{High-level trajectory estimation} Alignment over linguistic and visual sequences is formulated as a task of predicting a set of spans from the instruction that correspond to the current step. This process starts with a submodule $g_{PMTP}$ that estimates a count $cnt$ of steps from a high-level view on the route (see Figure \hyperref[pmtp]{4}). Path traces - denoted as ${tr}_T$ - are visual representations of trajectories generated from the coordinates of nodes. At $t_0$ in ${tr}_T$ initial spans in the instruction are assumed to align with the first visual perspective. From step $t_1$, a submodule containing a pretrained ConvNeXt Tiny model~\citep{liu2022convnet} updates an estimate of the step count in $cnt_{{tr}_T}$. A trajectory plan ${tp}_t$ is a Gaussian distribution of spans in $\uptau$ within the interval $[x_{left}, x_{right}]$. At each step, samples from this distribution serve as a prediction for relevant spans. The final output $\imath_t’$ is the predicted span $\imath_t$ combined with $\imath_{t-1}$.  

\begin{figure}[hbt!]
\begin{center}
\centerline{\includegraphics[width=\columnwidth]{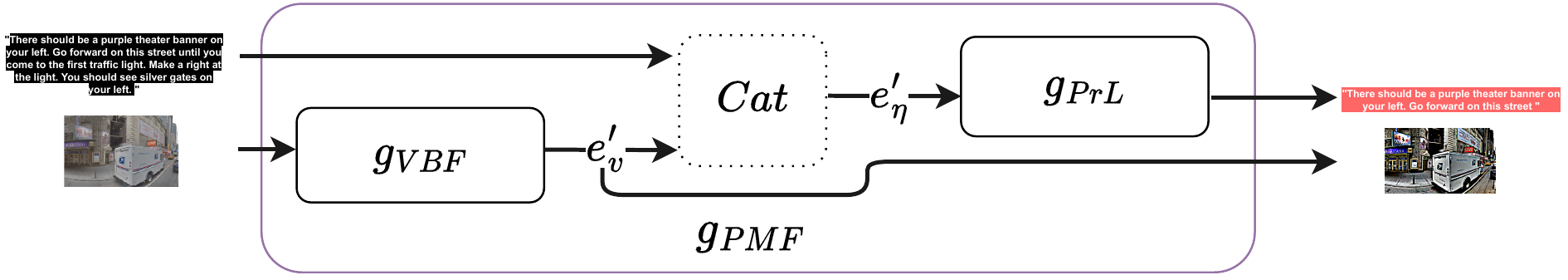}}
\caption{Submodule $g_{PMF}$ commences feature-level operations by boosting visual features in the perspective. The next operation ($Cat$) is a concatenation of the output from $g_{VBF}$ and the linguistic output $\imath_t’$ from the alignment process above. A precise prediction for the relevant span $e_l$ is returned by $g_{PrL}$.}
\label{pmf}
\end{center}
\vskip -0.3in
\end{figure}

\textbf{Feature-level localisation} 
Predicted spans are passed with $\psi_t$ to a submodule $g_{PMF}$ that is pretrained on cross-modal matching in $\phi_2$ (see Figure \hyperref[pmf]{5}). Feature-level operations commence with visual boost filtering. Let $Conv_{VBF}$ be a convolutional layer with a kernel $\kappa$ and weights $W$ that receives as input $\psi_t$. In the first operation $g_{USM}$, a Laplacian of Gaussian kernel $\kappa_{LoG}$ is applied to $\psi_t$. The second operation $g_{VBF}$ consists of subtracting the output $e_v$ from the original tensor $\psi_t$:
\begin{equation}
g_{VBF}(e_v) = (\lambda-1)(e_v)-g_{(USM)}(\psi_t)
\end{equation}
where $\lambda$ is a learned parameter for the degree of sharpening.

A combination of $g_{USM}$ and $g_{VBF}$ is equivalent to adaptive sharpening of details in an image with a Laplacian residual~\citep{bayer1986method}. Here operations are applied directly to $e_v$ and adjusted at each update of $W_{Conv_{VBF}}$ with a parameterised control $\beta\lambda$. In the simple and efficient implementation from \citet{carranza2019unsharp}, $\sigma$ in the distribution $LoG(\mu_j, \sigma_j)$ is fixed and the level of boosting is reduced to a single learned term
\begin{equation}
\Delta z(x_1, x_2) = \beta\lambda({\sum_j} ({AA}_{\kappa_{i_j}}' - A_{W_{\kappa_{i_j}}})_z) \newline
\end{equation}
where $A_W$ is a matrix of parameters and $AA'$ is the identity.

Selection of a localised span $e_l$ proceeds with a learned cross-modal embedding $e_{\eta}'$ composed of $e_v'$ and the linguistic output $\imath_t'$ from the preceding alignment operation. A binary prediction over this linguistic pair is performed on the output hidden state from a single-layer LSTM, which receives $e_{\eta}'$ as its input sequence. Function $g_{PrL}$ returns a precise localisation of relevant spans w.r.t. prominent features in the perspective:
\begin{equation}
\medmuskip=2mu   
\thickmuskip=3mu 
\renewcommand\arraystretch{1.5}
g_{PrL}(e_l) = g_{Cat}(\imath_t', e_v’)\triangleq 
\begin{cases}
  0, \, if \langle w, x \rangle + b < 0 \\    
  1, \, otherwise    
\end{cases}
\end{equation}

\textbf{Pretraining Strategy}
A data-efficient pretraining strategy for the PM-VLN module consists of pretraining submodules of the PM-VLN on auxiliary tasks $(\phi_1, \phi_2)$. We denote the two datasets for these tasks as $(\mathcal{D}_{\phi_1}, \mathcal{D}_{\phi_2})$ and a training partition as $\mathcal{D}^{Train}$ (see Appendix \ref{app_b} for details). In $\phi_1$, the $g_{PMTP}$ submodule is pretrained on TR-NY-PIT-central - a new set of path traces. Path traces in $D^{Train}_{\phi_1}$ are generated from 17,000 routes in central Pittsburgh with a class label for the step count in the route. The distribution of step counts in $D^{Train}_{\phi_1}$ is 50 samples for routes with $\le$7 steps and 300 samples for routes with $>$7 steps (see Appendix \ref{app_b}). During training, samples from $D^{Train}_{\phi_1}$ are presented in standard orientation for 20 epochs and rotated 180\textdegree{} ahead of a second round of training. This rotation policy is preferred following empirical evaluation using standalone versions of the $g_{PMTP}$ submodule receiving two alternate preparations of $D^{Train}_{\phi_1}$ with random and 180\textdegree{} rotations. Training is formulated as multiclass classification with cross-entropy loss on a set of M=66 classes
\begin{equation}
g_{\phi_1}(x_{tr}, \Theta) = B_0 + \underset{i}{argmax}\sum^M_{j=1}B_{i}(x_{tr}, W_j)
\end{equation}
where a class is the step count, $B$ is the bias, and $i$ is the sample in the dataset.

Pretraining on $\phi_2$ for the feature-level localisation submodule $g_{PMF}$ is conducted with the component integrated in the framework FL\textsubscript{PM} and the new MC-10 dataset. Samples in $D^{Train}_{\phi_2}$ consist of 8,100 landmarks in 10 US cities. To demonstrate the utility of open source tools in designing systems for outdoor VLN, the generation process leverages free and accessible resources that enable targeted querying. Entity IDs for landmarks sourced from the Wikidata Knowledge Graph are the basis for downloading textual summaries and images from the MediaWiki and WikiMedia APIs. Additional details on MC-10 are available in Appendix \ref{app_b}. The aim in generating the MC-10 dataset is to optimise $\Theta_{g_{PMF}}$ such that features relating to $Y_{\phi_{VLN}}$ are detected in inputs $X_{\phi_{VLN}}$. We opt for open  A multi-objective loss for $\phi_2$ consists of cross-modal matching over the paired samples $(x_l, x_v)$ - and a second objective comprising a prediction on the geolocation of the entity. In the first objective, $g_{PMF}$ conducts a binary classification between the true $x_l$ matching $x_v$ and a second textual input selected at random from entities in the mini-batch. A limit of 540 tokens is set for all textual inputs and the classification in $g_{PMF}$ is performed on the first sentence for each entity. Parameters $\Theta_{g_{PMF}}$ are saved and used subsequently for feature-level localisation in $\phi_{VLN}$.

\subsubsection{Cross-modal Attention and Action Prediction on Combined Outputs}
Resuming operations subsequent to the PM-VLN, outputs ${e_v’}_t$ from $Conv_{VBF}$ are passed together with ${e_l}_t$ to a VisualBERT embedding layer. Embeddings for both modalities are then processed by 4 transformer encoder layers with a hidden size of 256 and self-attention $\bigoplus$ is applied to learn alignments between the pairs
\begin{equation}
\label{eq:crosmodatt}
\medmuskip=2mu   
\thickmuskip=3mu 
\renewcommand\arraystretch{1.5}
\widetilde{e}_\eta  = \bigoplus(e_l \Longleftrightarrow e_v')  =  Soft\left(\sum^{\Ele}_{k=1}{\Msk}_k \text{\L}({\Ele}_k, \widetilde{\Ele}_k)\right) 
\end{equation}
where $Soft$ is the softmax function, $k$ is the number of elements in the inputs, $\Msk_{k=1}$ is a masked element over the cross-modal inputs, \L \ is the loss, $\Ele_k$ is an element in the input modality, and $\widetilde{\Ele}_k$ is the predicted element. Cross-modal embeddings resulting from this attention operation are processed by concatenating over layer outputs $g(\widetilde{{e}_\eta}’) = (\widetilde{e}_\Layer^1, \widetilde{e}_\Layer^2, \widetilde{e}_\Layer^3, \widetilde{e}_\Layer^4)$.

Architectural and embedding selections for our frameworks aim to enable comparison with benchmark systems on $\phi_{VLN}$. The $Enc_{Trans}$ in the best performing framework uses a standard VisualBERT encoder with a hidden size of 256 and 4 layers and attention heads. As noted above, inputs for $Enc_{Trans}$ align with those used in prior work~\citep{zhu2021multimodal}. 

A submodule $g_{CFns}$ combines $U_{\eta}$ from $\Layer^4$ of the $Enc_{Trans}$ and outputs from the cross-modal attention operation $g({\widetilde{E}_\eta}’)$ ahead of applying dropout. Predictions for navigation actions are the outputs of a classifier block consisting of linear layers with maxout activation. Maxout activation in a block composed of linear operations takes the $max z_{ij}$ where $z_{ij}$ are the product of $x_{ij}W_{n*}$ for $k$ layers. In contrast to ReLU, the activation function is learned and prevents unit saturation associated with performing dropout~\citep{goodfellow2013maxout}. We compare a standard classifier to one with $max \ x_i$ in Table \hyperref[table:ablations]{2}. Improvements with $max \ x_i$ are consistent with a requirement to offset variance when training with the high number of layers in the full FL\textsubscript{PM} framework.

\subsection{Theoretical Basis}
\label{theo}

This section provides a theoretical basis for a hierarchical process of cross-modal prioritisation that optimises attention over linguistic and visual inputs. In this section we use $q$ to denote this process for convenience. During the main task $\phi_{VLN}$, $q$ aligns elements in temporal sequences $\uptau$ and $Route$ and localises spans and visual features w.r.t. a subset of all entities $Ent$ in the routes:
\begin{equation}
q_{PM} = {\lVert x_l - x_v \rVert} {\underset{subject \, to} \rightarrow max} \; P_{D_{Ent}}[\uptau, Route] \leq R
\end{equation}

Inputs in $\phi_{VLN}$ consist of a linguistic sequence $\uptau$ and a visual sequence $Route$ or each trajectory $j$ in a set of trajectories. As a result of the process followed by \cite{chen2019touchdown} to create the Touchdown task, these inputs conform to the following definition.

\textbf{Definition 1} (Sequences refer to corresponding entities). \textit{At each step in $j$, $|x_l|$ and $|x_v|$ are finite subsequences drawn from $\uptau_j$ and $Route_j$ that refer to corresponding entities appearing in the trajectory $ent_j \subset Ent$.}

In order to simplify the notation, these subsequences are denoted in this section as $x_l$ and $x_v$. Touchdown differs from other outdoor navigation tasks~\citep{hermann2020learning} in excluding supervision on the alignment over cross-modal sequences. Furthermore $len(\uptau_j) \neq len(Route_j)$ and there are varying counts of subsequences and entities in trajectories. In an approach to $\phi_{VLN}$ formulated as supervised classification, an agent’s action at each step $\alpha_t \equiv$ classification $c_t \in \{0, 1\}$ where $c$ is based on corresponding $ent_t$ in the pair $(x_l, x_v)_t$. The likelihood that $c_t$ is the correct action depends in turn on detecting $S$ signal in the form of $ent_t$ from noise in the inputs. The objective of $q$ then is to maximise $P_S$ for each point in the data space.

The process defining $q$ is composed of multiple operations to perform two functions of high-level alignment $g_{Align}$ and localisation $g_{Loc}$. At the current stage $stg$, function $g_{Align}$ selects one set of spans $\varphi_{stg} \in (\varphi_1, \varphi_2, …, \varphi_n)$ \\ where $stg$ $\begin{cases}
  Start, \, \textit{if } t = 0 \\    
  End, \, \textit{if } t = -1 \\
  \forall \ stg_{other}, \, n \in N \in \sum_{n=1}^{n_1} > t_{-1} \ otherwise.    
\end{cases} \\ $
\\ This is followed by the function $g_{Loc}$, which predicts one of $\varsigma_{scnt_0} \lor \varsigma_{scnt_{0-1}}$ as the span $\varsigma$ relevant to the current trajectory step $scnt$  \\
 \\ where $scnt$ $\begin{cases}
  scnt_0, \, \textit{if } (\uptau, \psi_t) = 0 \\    
  scnt_{0-1}, \ otherwise.
 \end{cases} $

We start by describing the learning process when the agent in $\phi_{VLN}$ is a transformer-based architecture $Enc+Clas$ excluding an equivalent to $q$ (e.g. VisualBERT in Table 1 of the main report). $Enc+Clas$ is composed of two core subprocesses: cross-modal attention to generate representations $q(\bigoplus(L \Longleftrightarrow\widetilde V))$ and a subsequent classification $Clas(\widetilde{{e}_\eta}’)$. 

\textbf{Definition 2} (Objective in $Enc+Clas$). \textit{The objective $Obj_1(\theta)$ for algorithm $q(\bigoplus(L \Longleftrightarrow\widetilde V)$, where $L$ and $V$ are each sequences of samples $\{x_1, x_2, …, x_n\}$, is the correspondence between samples $x_l$ and $x_v$ presented at step $t$ in $\sum^n_{i=1}t_i = t_1 + t_2, … + t_n$.}

It is observed that in the learning process for $Enc+Clas$, any subprocesses to align and localise finite sequences $x_l$ and $x_v$ w.r.t. $ent_j$ are performed as implicit elements in the process of optimising $Obj_1(\theta)$. In contrast the basis for the hierarchical learning process enabled by our framework FL\textsubscript{PM} - which incorporates $q_{PM}$ with explicit functions for these steps - is given in Theorem 1.

\textbf{Theorem 1}. \textit{Assuming $x_l$ and $x_v$ conform to Definition 1 and that $\forall \ x \in L \ \exists \ x \in V$, an onto function $g_{Map} = mx + b, m \neq 0$ exists such that:
\begin{equation}
g_{Map}(x_l, x_v) \rightarrow max \left[ent^{(x_l, x_v)}_j \in Ent \right]
\end{equation}
In this case, additional functions $g_{Align}$ and $g_{Loc}$ - when implemented in order - maximise $g_{Map}$:
\begin{equation}
\begin{split}
max \ P_{D_{ent_j}} & = max \ g_{Map} (x_l, x_v) {\underset{subject \, to} \rightarrow} \\ & (\overrightarrow{g_{Align}, g_{Loc}, g_{Map}}) \ \forall \ ent^{(x_l, x_v)}_j \in L_j \cap V_j
\end{split}
\end{equation}}

\textbf{Remark 1} \textit{Let $P(max \ g_{Map})$ in Theorem 1 be the probability of optimising $g_{Map}$ such that the number of pairs $N^{(x_l, x_v)}$ corresponding to $ent_j \in L_j \cap V_j$ is maximised. It is noted that $N^{(x_l, x_v)}$ is determined by all possible outcomes in the set of cases $\{(x_l, x_v) \Leftrightarrow ent_j$, $(x_l, x_v) \nLeftrightarrow ent_j$, $x_l \nLeftrightarrow x_v$\}. As the sequences of instances $i$ in $x_l$, $x_v$ and $ent_j$ are forward-only, it is also noted that $N^{(x_l, x_v)}_{t+1} < N^{(x_l, x_v)}_{t}$ if $ent_i \not\in {x_l}_i$, $ent_i \not\in {x_v}_i$, or $ent^{x_l}_i \neq ent^{x_v}_i$. By definition, $N^{(x_l, x_v)}_{t+1} > N^{(x_l, x_v)}_{t}$ if $P(ent_i = {x_l}_i = {x_v}_i)$ - where the latter probability is s.t. processes performed within finite computational time $CT(n)$ - which implies that $P(max \ g_{Map}) | P(ent_i = {x_l}_i = {x_v}_i)$.}

\textbf{Remark 2}. \textit{Following on from Remark 1, \\ $CT(n^{P(ent_i = {x_l}_i = {x_v}_i)})$ \ when \ $q$ \ contains \ $g_t$, \ and \ function \\ $g_t(max(N^{(x_l, x_v)} \Rightarrow ent_j \in L_j \cap V_j), \ where \ g_t \in G < \\ CT(n^{P(ent_i = {x_l}_i = {x_v}_i)}) $ \ when \ $q$ \ does \ not \ contain \ $g_t < \\ CT(n^{P(ent_i = {x_l}_i = {x_v}_i)})$ \ when \ $q$ \ contains \ $g_t$, \ and \ function \\ $g_t(max(N^{(x_l, x_v)} \nRightarrow ent_j \in L_j \cap V_j)$.}

\textbf{Discussion}  In experiments, we expect from Remark 1 that results on $\phi_{VLN}$ for architectures such as $Enc+Clas$ - which exclude operations equivalent to those undertaken by the onto function $g_{Map}$ - will be lower than the results for a framework FL\textsubscript{PM} over a finite number of epochs. We observe this in Table \hyperref[table:res1]{1} when comparing the performance of respective standalone and + FL\textsubscript{PM}  for VisualBERT and VLN Transformer systems. Poor results for variants (a) and (h) in Tables \hyperref[table:ablations]{2} and \hyperref[table:trainstrategy]{3} in comparison to FL\textsubscript{PM} + VisualBERT(4l) also support the expectation set by Remark 2 that performance will be highly impacted in an architecture where operations in $g_{Map}$ increase the number of misalignments.

\textbf{Proof of Theorem 1}
\textit{We use below $a*$ for a generic transformer-based system that predicts $\alpha$ on $(L, V)$, $\nabla x$ for gradients, and $\Theta^{a*}$ to denote $\Theta^{Enc+Clas} \ \nu \ \Theta^{Enc + q}$.  
Let sequence $x_l = [ent_1, ent_2, …, ent_{n_1}]$ and sequence $x_v = [ent_1, ent_2, …, ent_{n_2}]$, where $n_1$ and $n_2$ are unknown. We note that at any point during learning, $P_S(x_l, x_v)$ is spread unevenly over $ent_j$ in relation to $\Theta^{a*} \approx \mathcal{X}$.}

\textbf{Propositions}
\textit{We start with the case that $\exists \ ent_j: ent^{(x_l)}$ $\ and \ ent^{(x_v)}$. Here \ $CT(n^{Ent \in L \cap V}) \ for \ \Theta^{a* + g_t} < CT(n^{Ent \in L \cap V}) \ for \ \Theta^{a*} \ where \ g_t$ accounts for $\Delta (Len_1, Len_2)$. We next consider the case where \ $\nexists \ ent_j: \ ent^{(x_l)} \ \nu \ ent^{(x_v)}$. $Where \ \nexists \ g_{Loc} \ then \ P_S^{(x_l, x_v)} < \exists \ g_{Loc} \ P_S^{(x_l, x_v)}$. \\ We conclude with the case where $\exists \ Ent: x_l \ \nu \ x_v$. \ $In \ P_S^{A*} \ ent^{(x_l)} \ \bigoplus \ ent^{(x_v)} \ when \ ent^{(x_l)} \ \neq \ ent^{(x_v)}.$}

\textit{As} \ $(Ent_L, Ent_V) \ \Rightarrow \ Ent$, $\Theta^{a*} \ \approx \ max(N^{(x_l, x_v)}) \ \in \ \mathcal{X}.$ \\ \ $P_S^{(x_l, x_v)} \ where \ ent_i \ = \ {x_l}_i \ = \ {x_v}_i \ > \ ent_i \ \in \ \Theta^{a*} \ \approx \\ max(N^{(x_l, x_v)}).$ \textit{Furthermore} $P \ \exists \ ent \in \ Ent \approx (ent_i) \ > \nexists \\ \ ent \ \not\approx \ ent_i.$ \textit{Therefore} \ $slope \ \nabla x \ increases \ and \ CT(n^{Ent \in L \cap V}) \\ for \ \Theta^{a* + q} < CT(n^{Ent \in L \cap V}).$ 


\begin{table*}[hbt!]
\label{table:res1}
\begin{center}
\begin{threeparttable}
\begin{tabular}{l l l l l l l l}
\toprule
                                &               & \multicolumn{3}{l}{\textbf{Development}} & \multicolumn{3}{l}{\textbf{Test}} \\ 
                                &            & \textbf{TC$\uparrow$}    & \textbf{SPD$\downarrow$}   & \textbf{SED$\uparrow$}   & \textbf{TC$\uparrow$}   & \textbf{SPD$\downarrow$}   &    \textbf{SED$\uparrow$}\\ \midrule
Inputs (L, V)   & GA \tnote{a} & 12.1    & 20.2   & 11.7   & 10.7   & 19.9   & 10.4   \\
(non-transformer based)          & RCONCAT \tnote{a} & 11.9    & 20.1   & 11.5   & 11.0   & 20.4   & 10.5   \\ 
                                & ARC+L2STOP* \tnote{c}  & 19.5    & 17.1   & 19.0   & \textbf{16.7}   & 18.8   & 16.3   \\ 
                                 \cmidrule{1-8}
Inputs (L, V)                   & VisualBERT(8l)  & 10.4    & 21.3   & 10.0   & 9.9   & 21.7   & 9.5 \\  
(transformer based)             & VisualBERT(4l)  & 14.3    & 17.7   & 13.7   & 11.8   & 18.3   & 11.5   \\
               & VLN Transformer(4l) \tnote{b}  & 12.2    & 18.9   & 12.0   & 12.8   & 20.4   & 11.8   \\
               & VLN Transformer(8l) \tnote{b}  & 13.2    & 19.8   & 12.7   & 13.1   & 21.1   & 12.3   \\
                & VLN Transformer(8l) + M50 + style * \tnote{b}  & 15.0    & 20.3   & 14.7   & \textbf{16.2}   & 20.8   & 15.7   \\
                                \cmidrule{1-8}
Inputs (L, V) + JD / HT**      & ORAR (ResNet pre-final)* \tnote{d}  & 26.0    & 15.0   & -   & 25.3   & 16.2   & -   \\
(non-transformer based) & ORAR (ResNet 4th-to-last)* \tnote{d}  & 29.9    & 11.1   & -   & \textbf{29.1}   & 11.7   & -   \\

                                \cmidrule{1-8}
                               
Inputs (L, V) + Path Traces      & VLN Transformer(8l)  & 11.2    & 23.4   & 10.7   & 11.5   & 23.9                                   &                                            10.8   \\ 
 (transformer based)   & VisualBERT(4l)  & 16.2    & 18.7   & 15.7   & 15.0   & 20.1                                   &                                            14.5   \\
                                & FL\textsubscript{PM}(4l) $+$ VLN Transformer(8l)  & 29.9    & 23.4   & 26.8   & 28.2   & 23.8   & 25.6 \\
                                & FL\textsubscript{PM}(4l) $+$ VisualBERT(4l)  & 33.0    & 23.6   & 29.5   & \textbf{33.4}   & 23.8   & 29.7   \\\bottomrule
\end{tabular}
\begin{tablenotes}
\item Frameworks from \textsuperscript{a} \cite{chen2019touchdown},
\textsuperscript{b} \cite{zhu2021multimodal},
\textsuperscript{c} \cite{xiang-etal-2020-learning}, and
\textsuperscript{d} \cite{schumann-riezler-2022-analyzing}.
\item[*] Results reported by the authors.
\item[**] Systems receive two types of features - Junction Type and Heading Delta - as inputs.
\end{tablenotes}
\caption{Performance on the Touchdown benchmark ranked by TC on the test partition. Systems are grouped by input types during VLN and the use of transformer blocks in architectures. Contributions of the  FL\textsubscript{PM} framework and path traces to improved performance are demonstrated with results for systems with two baseline transformer-based architectures - VisualBERT and VLN Transformer. These baselines also are assessed in two sizes to test the benefits of adding transformer blocks.}
\end{threeparttable}
\end{center}
\vspace{-5mm} 
\end{table*}

\section{Experiments}

Our starting point in evaluating the PM-VN module and FL\textsubscript{PM} is performance in relation to benchmark systems (see Table \hyperref[table:res1]{1}). Ablations are conducted by removing individual operations (see Table \hyperref[table:res1]{2}) and the role of training data is assessed (see Table \hyperref[table:res1]{3}). To minimise computational cost, we implement frameworks with low numbers of layers and attention heads in transformer models.

\subsection{Experiment Settings}
\textbf{Metrics \ } 
We align with~\citet{chen2019touchdown} in reporting task completion (TC), shortest-path distance (SPD), and success weighted edit distance (SED) for $\phi_{VLN}$. All metrics are derived using the Touchdown navigation graph. TC is a binary measure of success ${0, 1}$ in ending a route with a prediction $c_{t-1}^\omicron = y_{t-1}^\omicron$ or $c_{t-1}^\omicron = y_{t-1}^{\omicron-1}$ and SPD is calculated as the mean distance between $c_{t-1}^\omicron$ and $y_{t-1}^\omicron$. SED is the Levenshtein distance between the predicted path in relation to the defined route path and is only applied when TC = 1.

\textbf{Hyperparameter Settings \ } 
Frameworks are trained for 80 epochs with batch size=30. Scores are reported for the epoch with the highest SPD on $\mathcal{D}_{\phi_{VLN}}^{Dev}$. Pretraining for the PM-VLN module is conducted for 10 epochs with batch sizes $\phi_1=60$ and $\phi_2=30$. Frameworks are optimised using AdamW with a learning rate of 2.5 x 10\textsuperscript{-3}~\citep{loshchilov2017decoupled}.

\subsection{Touchdown}

\textbf{Experiment Design:} 
\cite{chen2019touchdown} define two separate tasks in the Touchdown benchmark: VLN and spatial description resolution. This research aligns with other studies~\citep{zhu2021multimodal, zhu-etal-2022-diagnosing} in conducting evaluation on the navigation component as a standalone task. \textbf{Dataset and Data Preprocessing:} Frameworks are evaluated on full partitions of Touchdown with $D^{Train}=6,525$, $D^{Dev}=1,391$, and $D^{Test}=1,409$ routes. Trajectory lengths vary with $D^{Train}=34.2$, $D^{Dev}=34.1$, and $D^{Test}=34.4$ mean steps per route. Junction Type and Heading Delta are additional inputs generated from the environment graph and low-level visual features~\citep{schumann-riezler-2022-analyzing}.  M-50 + style is a subset of the StreetLearn dataset with $D^{Train}=30,968$ routes of 50 nodes or less and multimodal style transfer applied to instructions~\citep{zhu2021multimodal}. \textbf{Embeddings:} All architectures evaluated in this research receive the same base cross-modal embeddings $x_{\eta}$ proposed by \cite{zhu2021multimodal}, which are learned by a combination of the outputs of a pretrained BERT-base encoder with 12 encoder layers and attention heads. At each step, a fully connected layer is used for textual embeddings $\varsigma_t$ and a 3 layer CNN returns the perspective $\psi_t$. FL\textsubscript{PM} frameworks also receive an embedding of the path trace $tr_t$ at step $t$. As this constitutes additional signal on the route, we evaluate a VisualBERT model (4l) that also receives $tr_t$, which in this case is combined with $\psi_t$ ahead of inclusion in ${x_{\eta}}_t$. \textbf{Results:} In Table \hyperref[table:res1]{1} the first block of frameworks consists of architectures composed primarily of convolutional and recurrent layers. VLN Transformer is a framework proposed by \cite{zhu2021multimodal} for the Touchdown benchmark and consists of a transformer-based cross-modal encoder with 8 encoder layers and 8 attention heads. VLN Transformer + M50 + style is a version of this framework pretrained on the dataset described above. To our knowledge, this was the transformer-based framework with the highest TC on Touchdown preceding our work. ORAR (ResNet 4th-to-last)~\citep{schumann-riezler-2022-analyzing} is from work published shortly before the completion of this research and uses two types of features to attain highest TC in prior work. Standalone VisualBERT models are evaluated in two versions with 4 and 8 layers and attention heads. A stronger performance by the smaller version  indicates that adding self-attention layers is unlikely to improve VLN predictions. This is further supported by the closely matched results for the VLN Transformer(4l) and VLN Transformer(8l). FL\textsubscript{PM} frameworks incorporate the PM-VLN module pretrained on auxiliary tasks $(\phi_1, \phi_2)$ - and one of VisualBERT (4l) or VLN Transformer(8l) as the main model. Performance on TC for both of these transformer models doubles when integrated into the framework. A comparison of results for standalone VisualBERT and VLN Transformer systems with path traces supports the use of specific architectural components that can exploit this additional input type. Lower SPD for systems run with the FL\textsubscript{PM} framework reflect a higher number of routes where a stop action was predicted prior to route completion. Although not a focus for the current research, this shortcoming in VLN benchmarks has been addressed in other work~\citep{xiang-etal-2020-learning, blukis2018mapping}.

\subsection{Assessment of Specific Operations}

Ablations are conducted on the framework with the highest TC i.e. FL\textsubscript{PM} + VisualBERT(4l). The tests do not provide a direct measure of operations as subsequent computations in forward and backward passes by retained components are not accounted for. Results indicate that initial alignment is critical to cross-modal prioritisation and support the use of in-domain data during pretraining.

\begin{table}[hbt!]
\label{table:ablations}
\centering
\begin{tabular}{p{4cm} p{.8cm} p{.8cm} p{.8cm}}
\toprule
                                &                \multicolumn{3}{l}{\textbf{Development}}  \\ 
                                & \textbf{TC$\uparrow$}    & \textbf{SPD$\downarrow$}   & \textbf{SED$\uparrow$}\\ \midrule
FL\textsubscript{PM} $+$ VisualBERT(4l) & 33.0    & 23.6   & 29.5  \\ \cmidrule{1-4}
PM-VLN \\ \hspace{1mm} - $g_{PMTP}$ (a) & 7.1    & 26.8   & 6.8  \\ 
\hspace{1mm} - $g_{PMF}$ minus g\textsubscript{VBF} (b) & 27.9    & 25.7   & 24.9  \\
\hspace{1mm} - $g_{PMF}$ minus $\imath_{t-1}$  (c)  & 29.8    & 21.8   & 27.2 \\ \cmidrule{1-4}
FL\textsubscript{PM} \\
\hspace{1mm} - $g_{Attn}$ with $g_{Cat}$ (d) & 18.8    & 30.5   & 16.4  \\ 
\hspace{1mm} - $g_{Clas_{max \ x_i}}$ with $g_{Clas}$ (e) & 31.7    & 21.9   & 28.2 
                                \\\bottomrule
\end{tabular}
\caption{Ablations on core operations in the PM-VLN (variants (a-(c)) and the FL\textsubscript{PM} framework (variants (d) and (e)).}
\end{table}

\textbf{Ablation 1: PM-VLN \ }
Prioritisation in the PM-VLN module constitutes a sequential chain of operations. Table \hyperref[table:ablations]{2} reports results for variants of the framework where the PM-VLN excludes individual operations. Starting with $g_{PMTP}$, trajectory estimation is replaced with a fixed count of 34 steps for each route $tr_t$ (see variant (a)). This deprives the PM-VLN of a method to take account of the current route when synchronising $\uptau$ and sequences of visual inputs. All subsequent operations are impacted and the variant reports low scores for all metrics. Two experiments are then conducted on $g_{PMF}$. In variant (b), visual boost filtering is disabled and feature-level localisation relies on a base $\psi_t$. A variant excluding linguistic components from $g_{PMF}$ is then implemented by specifying $\imath_t$ as the default input from $\uptau_t$ (see variant (c)). In practice, span selection in this case is based on trajectory estimation only. 

\textbf{Ablation 2: FL\textsubscript{PM} \ }
Ablations conclude with variants of FL\textsubscript{PM} where core functions are excluded from other submodules in the framework. Results for variant (d) demonstrate the impact of replacing the operation defined in Equation \hyperref[eq:crosmodatt]{3} with a simple concatenation on outputs from PM-VLN $e_l$ and $e_v'$. A final experiment compares methods for generating action predictions: in variant (e), $g_{Clas_{max \ x_i}}$ is replaced by the standard implementation for classification in VisualBERT. Classification with dropout and a single linear layer underperforms our proposal by 1.3 points on TC.

\subsection{Assessment of Training Strategy}

A final set of experiments is conducted to measure the impact of training data for auxiliary tasks $(\phi_1, \phi_2)$.

\textbf{Training Strategy 1: Exploiting Street Pattern in Trajectory Estimation \ }
We conduct tests on alternate samples to examine the impact of route types in $D^{Train}_{\phi_1}$. The module for FL\textsubscript{PM} frameworks in Table \hyperref[table:res1]{1} is trained on path traces drawn from an area in central Pittsburgh (see \textbf{SupMat:Sec.3}) with a rectangular street pattern that aligns with the urban grid type~\cite{lynch1981theory} found in the location of routes in Touchdown. Table \hyperref[table:trainstrategy]{3} presents results for modules trained on routes selected at random outside of this area. In variants (f) and (g), versions V2 and V3 of $D^{Train}_{\phi_1}$ each consist of 17,000 samples drawn at random from the remainder of a total set of 70,000 routes. Routes that conform to curvilinear grid types are observable in outer areas of Pittsburgh. Lower TC for these variants prompts consideration of street patterns when generating path traces. A variant (h) where the $g_{PMTP}$ submodule receives no pretraining underlines - along with variant (a) in Table \hyperref[table:ablations]{2} - the importance of the initial alignment step to our proposed method of cross-modal prioritisation.  

\textbf{Training Strategy 2: In-domain Data and Feature-Level Localisation \ }
We conclude by examining the use of in-domain data when pretraining the $g_{PMF}$ submodule ahead of feature-level localisation operations in the PM-VLN. In Table \hyperref[table:trainstrategy]{3}, versions of FL\textsubscript{PM} are evaluated subsequent to pretraining with varying sized subsets of the Conceptual Captions dataset~\cite{sharma2018conceptual}. This resource of general image-text pairs is selected as it has been proposed for pretraining VLN systems (see \hyperref[pretrainrl]{below}). Samples are selected at random and grouped into two training partitions equivalent in number to $100\%$ (variant(i)) and $150\%$ of $D^{Train}_{\phi_2}$ (variant (j)). In place of the multi-objective loss applied to the MC-10 dataset, $\theta_{g_{PMF}}$ are optimised on a single goal of cross-modal matching. Variant (k) assesses FL\textsubscript{PM} when no pretraining for the $g_{PMF}$ submodule is undertaken. Lower results for variants (i), (j), and (k) support pretraining on small sets of in-domain data as an alternative to optimising VLN systems on large-scale datasets of general samples.

\begin{table}[hbt!]
\label{table:trainstrategy}

\begin{tabular}{p{4cm} p{.8cm} p{.8cm} p{.8cm}}
\toprule
                                &                \multicolumn{3}{l}{\textbf{Development}}  \\ 
                                & \textbf{TC$\uparrow$}    & \textbf{SPD$\downarrow$}   & \textbf{SED$\uparrow$}\\ \midrule
FL\textsubscript{PM} $+$ VisualBERT(4l) & 33.0    & 23.6   & 29.5   \\ \cmidrule{1-4}
Pretraining for $g_{PMTP}$ \\
\hspace{1mm} - $g_{PMTP} + D^{Train}_{\phi_1} V2$ (f) & 11.9    & 20.1   & 11.5  \\ 
\hspace{1mm} - $g_{PMTP} + D^{Train}_{\phi_1} V3$ (g) & 13.6    & 20.5   & 13.1
\\
\hspace{1mm} - $g_{PMTP}$ no pretraining (h) & 4.7    & 27.6   & 1.9
\\ \cmidrule{1-4}
Pretraining for $g_{PMF}$ \\
\hspace{1mm} - $g_{PMF} + D^{Train}_{\phi_2} V2$ (i) & 19.8    & 23.2   & 17.2 \\
\hspace{1mm} - $g_{PMF} + D^{Train}_{\phi_2} V3$ (j) & 23.9    & 20.8   & 20.3 \\
\hspace{1mm} - $g_{PMF}$ no pretraining (k) & 6.3    & 25.1   & 4.6
                                \\\bottomrule
\end{tabular}
\caption{Assessment of the pretraining strategy for individual PM-VLN submodules $g_{PMTP}$ (variants (f) to (h)) and $g_{PMF}$ (variants (i) to (k) using alternative datasets for auxiliary tasks. Variants are also run with no pretraining of $g_{PMTP}$ and $g_{PMF}$.}
\end{table}

\section{Related Work} 
This research aims to extend cross-disciplinary links between machine learning and computational cognitive neuroscience in the study of prioritisation in attention. This section starts with a summary of literature in these two disciplines that use computational methods to explore this subject. Our training strategy is positioned in the context of prior work on pretraining frameworks for VLN. The section concludes with work related to the alignment and feature-level operations performed by our PM-VLN model. 

\textbf{Computational Implementations of Prioritisation in Attention}
\citet{denil2012learning} proposed a model that generates saliency maps where feature selection is dependent on high-level signals in the task. The full system was evaluated on computer vision tasks where the aim is to track targets in video. A priority map computation was implemented in object detection models by \citet{wei2016learned} to compare functions in these systems to those observed in human visual attention. \citet{anantrasirichai2017fixation} used a Support Vector Machine classifier to model visual attention in human participants traversing four terrains. Priority maps were then generated to study the interaction of prioritised features and a high-level goal of maintaining smooth locomotion. A priority map component was incorporated into a CNN-based model of primate attention mechanisms by \cite{zelinsky2019learning} to prioritise locations containing classes of interest when performing visual search. Studies on spatial attention in human participants have explored priority map mechanisms that process inputs consisting of auditory stimuli and combined linguistic and visual information~\citep{golob2017computational, cavicchio2014effect}. To our knowledge, our work is the first to extend neuropsychological work on prioritisation over multiple modalities to a computational implementation of a cross-modal priority map for machine learning tasks.
 
\textbf{Pretraining for VLN Tasks}
Two forms of data samples - in-domain and generic - are used in pretraining prior to conducting VLN tasks. In-domain data samples have been sourced from image-caption pairs from online rental listings~\citep{guhur2021airbert} and other VLN tasks~\citep{zhu2021multimodal}. In-domain samples have also been generated by augmenting or reusing in-task data~\citep{fried2018speaker, huang2019transferable, hao2020towards, he2021landmark, schumann-riezler-2022-analyzing}. Generic samples from large-scale datasets designed for other Vision-Language tasks have been sourced to improve generalisation in transformer-based VLN agents. \citet{majumdar2020improving} conduct large-scale pretraining with 3.3M image-text pairs from Conceptual Captions \citet{sharma2018conceptual} and \citet{qi2021road} initialise a framework with weights trained on four out-of-domain datasets. Our training strategy relies on datasets with a few thousand samples derived from resources where additional samples are available at low cost.

\textbf{Methods for Aligning and Localising Features in Linguistic and Visual Sequences}
Alignment in multimodal tasks is often posited as an implicit subprocess in an attention-based component of a transformer~\citep{tsai2019multimodal, zhu2021multimodal}. \citet{huang2019transferable} identified explicit cross-modal alignment as an auxiliary task that improves agent performance in VLN. Alignment in this case is measured as a similarity score on inputs from the main task. In contrast, our PM-VLN module conducts a hierarchical process of trajectory planning and learned localisation to pair inputs. A similarity measure was the basis for an alignment step in the Vision-Language Pretraining framework proposed by \citet{li2021align}. A fundamental point of difference with our work is that this framework - along with related methods~\citep{jia2021scaling} - is trained on a distinct class of tasks where the visual input is a single image as opposed to a temporal sequence. Several VLN frameworks containing components that perform feature localisation on visual inputs have been pretrained on object detection~\citep{majumdar2020improving, suglia2021embodied, hu-etal-2019-looking}. In contrast, we include visual boost filtering in $g_{PMF}$ to prioritise visual features. Our method of localising spans using a concatenation of the enhanced visual input and cross-modal embeddings is unique to this research.

\section{Conclusion} 
We take inspiration from a mechanism described in neurophysiological research with the introduction of a priority map module that combines temporal sequence alignment enabled by high-level trajectory estimation and feature-level localisation. Two new resources comprised of in-domain samples and a tailored training strategy are proposed to enable data-efficient pretraining of the PM-VLN module ahead of the main VLN task. A novel framework enables action prediction with maxout activation on a combination of the outputs from the PM-VLN module and a transformer-based encoder. Evaluations demonstrate that our module, framework, and pretraining strategy double the performance of standalone transformers in outdoor VLN.

\section{Acknowledgments}

This publication is supported by the Digital Visual Studies program at the University of Zurich and funded by the Max Planck Society. RS has received funding by the Swiss National Science Foundation (project MUTAMUR; no.\ 176727). The authors would like to thank Howard Chen, Piotr Mirowski, and Wanrong Zhu for assistance and feedback on questions related to the Touchdown task, data, and framework evaluations.


{\footnotesize
\bibliography{main.bib}}

\newpage
\appendix
\begin{appendices}
\onecolumn
\begin{center}
\textbf{\huge Appendices}
\end{center}
\vspace{0.25in}
\section{Notation}
\label{app_a}
Notations used in multiple sections of this paper are defined here for fast reference. Auxiliary tasks $(\phi_1, \phi_2)$ and the main VLN task $\phi_{VLN}$ constitute the set of tasks $\Phi$. Inputs and embeddings are specified as $l$ (linguistic), $v$ (visual), and $\eta$ (multimodal). A complete textual instruction is denoted as $\uptau$, $\varsigma$ is a span, and $\psi$ is a perspective. Linguistic and visual inputs for the PM-VLN are denoted as $(\imath_t', \psi_t)$ and embeddings processed in prioritisation operations are $(e_l, e_v)_t$. In contrast, $U$ denotes a set of embeddings from the main model, which are derived from inputs $\bar{e}_{\eta}, \psi_{cat})$. The notations $\Delta$ and $\bigoplus$ are respectively visual boost filtering and self-attention operations. Table \hyperref[table:notation]{4} provides a reference source for standard notation appearing throughout this paper. Other notations are defined in the sections where they are used.

\begin{table}[hbt!] 
\label{table:notation}
\begin{center}
\begin{tabular}{l c c c}
\hline
\multicolumn{1}{c}{\textbf{Notation}}    &  \multicolumn{1}{c}{\textbf{Usage in this paper}} \\ \cline{1-2}
$A$ & Matrix \\
$AA$ & Identity matrix \\
$B, b$ & Bias \\
$\mathcal{D}$ & Dataset  \\
${Train, Dev, Test}$ & Dataset partitions \\
$\exists$ & Exists \\
$\forall$ & For every (eg member in a set) \\
$g$ & Function \\
$H$ & Hypothesis \\
$\Layer$ & Layer of a model \\
$len$ & Length \\
$\mu$ & Mean \\
$n$ & Number of samples \\
$\nu$ & Or \\
$P$ & Probability \\
$q$ & Algorithm \\
$S$ & Signal detected \\
$\sigma$ & Standard deviation \\
$\Theta$ & Set of parameters \\
$W, w$ & Set of weights \\
$|x|$ & Sequence \\
$\triangleq$ & Equal by definition \\
  \hline
\end{tabular}
\caption{Reference List for Standard Notation.}
\end{center}
\end{table}

\section{Datasets}
\label{app_b}
\subsection{Generation and Partition Sizes}
The MC-10 dataset consists of visual, textual and geospatial data for landmarks in 10 US cities. We generate the dataset with a modified version of the process outlined by \citet{armitage2020mlm}. Two base entity IDs - Q2221906 (``geographic location'') and Q83620 (``thoroughfare'') - form the basis of queries to extract entities at a distance of $<=2$ hops in the Wikidata knowledge graph\footnote{\url{https://query.wikidata.org/}}. Constituent cities consist of incorporated places exceeding 1 million people ranked by population density based on data for April 1, 2020 from the US Census Bureau\footnote{\url{https://www.census.gov/programs-surveys/decennial-census/data/datasets.html}}.  Images and coordinates are sourced from Wikimedia and text summaries are extracted with the MediaWiki API. Geographical cells are generated using the S2 Geometry Library\footnote{\url{https://code.google.com/archive/p/s2-geometry-library/}} with a range of $n$ entities $[1, 5]$. Statistics for MC-10 are presented by partition in Table \hyperref[table:mc10]{5}. As noted above, only a portion of textual inputs are used in pretraining and experiments.

\begin{table}[hbt!] 
\label{table:mc10}
\begin{center}
\caption{Statistics for the MC-10 dataset by partition.}
\vspace{-1em}
\begin{tabular}{l c c c c}
\hline
                                &  \multicolumn{1}{c}{\textbf{Train}} & \multicolumn{1}{c}{\textbf{Development}} &  \\ \cline{2-3} 
{\textbf{Number of entities}}  & 8,100    & 955   \\ 
 
{\textbf{Mean length per text summary}}  & 727    & 745  \\ 
  \hline
\end{tabular}
\end{center}
\end{table}

TR-NY-PIT-central is a set of image files graphing path traces for trajectory plan estimation in two urban areas. Trajectories in central Manhattan are generated from routes in the Touchdown instructions~\cite{chen2019touchdown}. Links $E$ connecting $\Omicron$ in the Pittsburgh partition of StreetLearn~\cite{mirowski2018learning} are the basis for routes where at least one node is positioned in the bounding box delimited by the WGS84 coordinates (40° 27' 38.82", -80° 1' 47.85") and (40° 26' 7.31", -79° 59' 12.86"). Labels are defined by step count $cnt$ in the route. Total trajectories sum to 9,325 in central Manhattan and 17,750 in Pittsburgh. In the latter location, routes are generated for all nodes with 50 samples randomly selected where $cnt =< 7$ and 200 samples where $cnt > 7$. The decision to generate a lower number of samples for shorter routes was determined by initial tests with the ConvNeXt Tiny model~\cite{liu2022convnet}. We opt for a maximum $cnt$ of 66 steps to align with the longest route in the training partition of Touchdown. The resulting training partition of samples for Pittsburgh consists of 17,000 samples and is the resource used to pretrain $g_{PMTP}$ in the PM-VLN module. 

\subsection{Samples from Datasets}

\begin{figure}[h!]
\begin{center}
\includegraphics[scale=0.4]{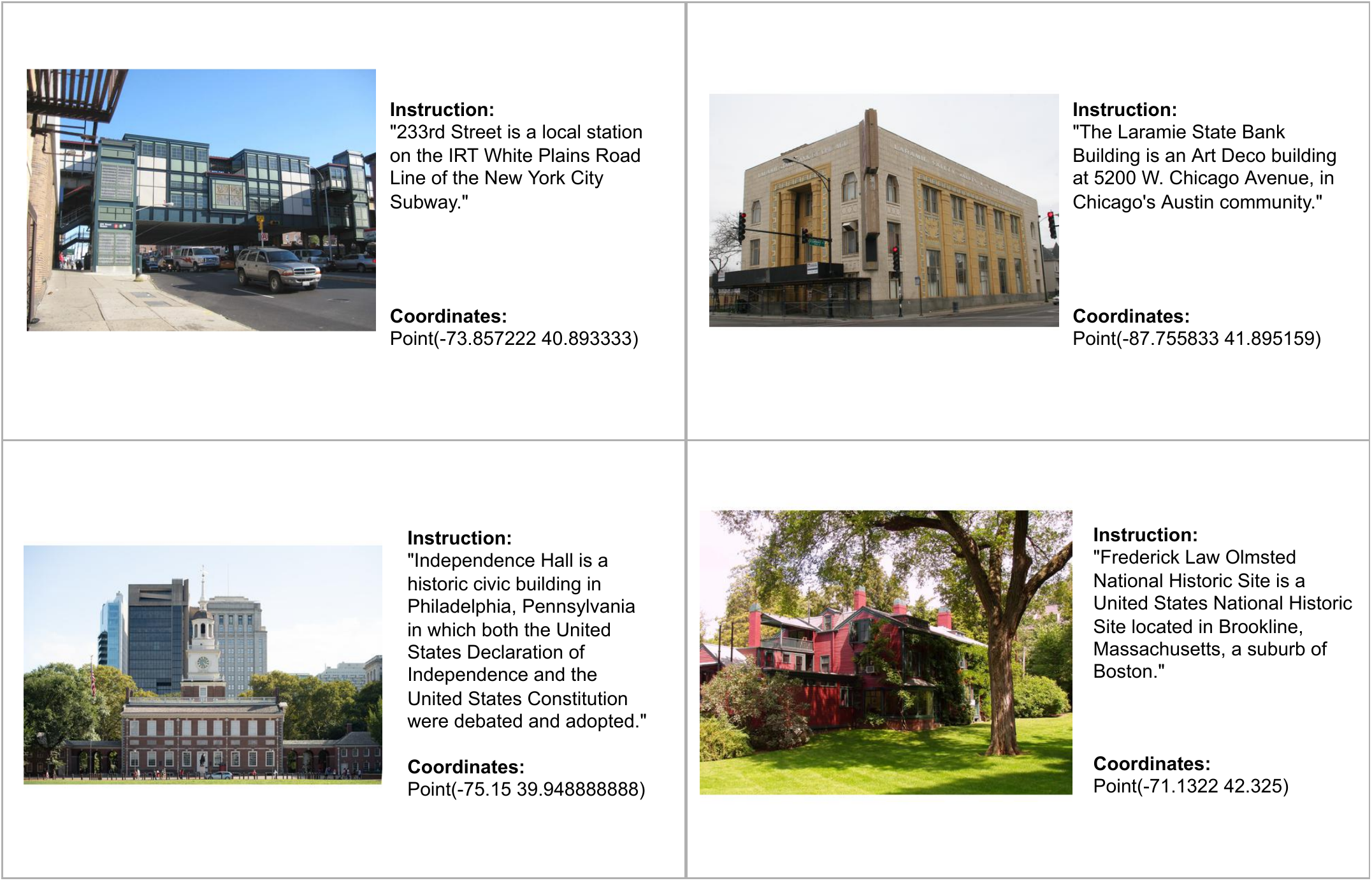}
\caption{Samples from the MC-10 dataset.}
\label{mc10sample}
\end{center}
\vskip -0.3in
\end{figure}

\begin{figure}[H]
\begin{center}
\includegraphics[scale=0.4]{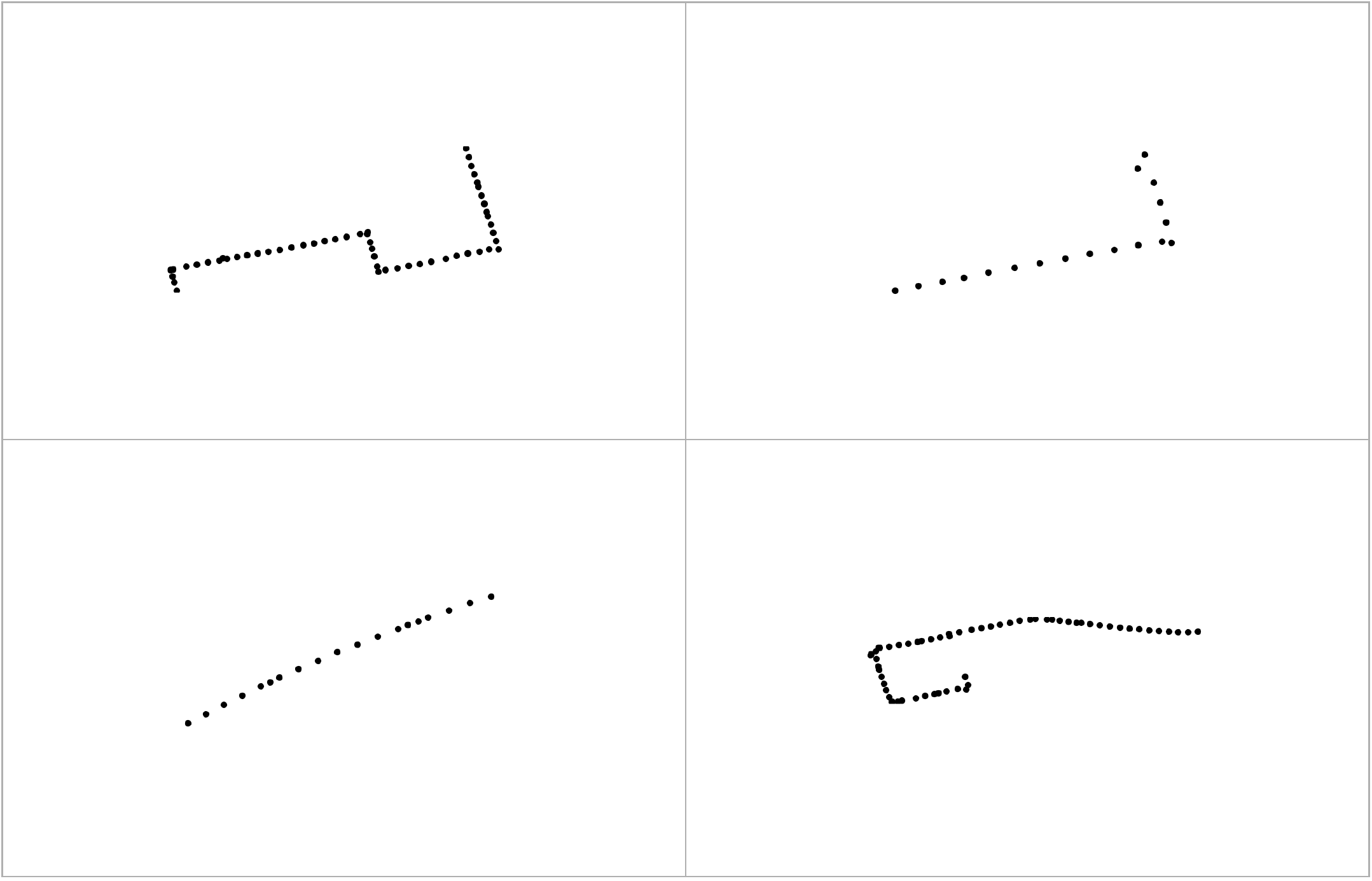}
\caption{Samples from the TR-NY-PIT-central dataset with path traces representing routes in central Pittsburgh.}
\label{trsample}
\end{center}
\vskip -0.3in
\end{figure}

In auxiliary task $\phi_2$, the $g_{PMF}$ submodule of PM-VLN is trained on visual, textual, and geodetic position data types.  Path traces from the TR-NY-PIT-central are used in $\phi_1$ to pretrain the $g_{PMTP}$ submodule on trajectory estimation. Samples for entities in MC-10 and path traces in TR-NY-PIT-central are presented in Figures \hyperref[mc10sample]{1} and \hyperref[trsample]{2}. 

\section{Code and Data}
\label{app_c}
Source code for the project and instructions to run the framework are released and maintained in a public GitHub repository under MIT license (\url{https://github.com/JasonArmitage-res/PM-VLN}). Code for the environment, navigation, and training adheres to the codebases released by \cite{zhu2021multimodal} and \cite{chen2019touchdown} with the aim of enabling comparisons with benchmarks introduced in prior work on Touchdown. Full versions of the MC-10 and TR-NY-PIT-central datasets are published on Zenodo under Creative Commons public license (\url{https://zenodo.org/record/6891965#.YtwoS3ZBxD8}).

\end{appendices}

\end{document}